# Construction of a Syntactic Analysis Map for Yi Shui School through Text Mining and Natural Language Processing Research

Hanqing ZHAO[1] * Yuehan LI[1]

(1.College of Traditional Chinese Medicine, Hebei University, Baoding 071000, China)

**Abstract:** Entity and relationship extraction is a crucial component in natural language processing tasks such as knowledge graph construction, question answering system design, and semantic analysis. Most of the information of the Yishui school of traditional Chinese Medicine (TCM) is stored in the form of unstructured classical Chinese text. The key information extraction of TCM texts plays an important role in mining and studying the academic schools of TCM. In order to solve these problems efficiently using artificial intelligence methods, this study constructs a word segmentation and entity relationship extraction model based on conditional random fields under the framework of natural language processing technology to identify and extract the entity relationship of traditional Chinese medicine texts, and uses the common weighting technology of TF-IDF information retrieval and data mining to extract important key entity information in different ancient books. The dependency syntactic parser based on neural network is used to analyze the grammatical relationship between entities in each ancient book article, and it is represented as a tree structure visualization, which lays the foundation for the next construction of the knowledge graph of Yishui school and the use of artificial intelligence methods to carry out the research of TCM academic schools.

## 1 Introduction

In the era of artificial intelligence and big data technology, the mining and utilization of ancient Chinese medicine literature knowledge is one of the important basic tasks for the inheritance and innovation and development of traditional Chinese medicine. With the progress of technology, although certain achievements[1] have

---

*Funded by National Natural Science Foundation of China (No.82004503) and Science and Technology Project of Hebei Education Department(BJK2024108)

Corresponding author: Hanqing Zhao, zhaohq@hbu.edu.cn

been made in related fields in recent years, there are still great challenges, especially in the inheritance and development of traditional Chinese medicine schools.

Most of the academic schools of TCM have been inherited in the form of ancient books, and the data are mainly in the form of unstructured text. The manual processing and extraction of ancient book data such as named entities i s time-consuming and labor-intensive. Ancient documents are recorded in classi cal Chinese, which uses concise words and words, and is quite different from modern Chinese in vocabulary and semantics. In particular, there is a lack of s tandard data sets for artificial intelligence analysis, which provides great obstacl es for computer methods to automatically extract ancient documents. At present, Chinese named entity recognition methods are mainly based on rule-based, sta tistical machine learning and deep learning methods[2]. Among them, the rule-ba sed method relies on manual rules, combines the named entity library, and dete rmines the type of the entity by the consistency between the entity and the rul es. This method can achieve good recognition results, but the rules in different fields are different and these rules cannot be used interactively. Therefore, ma chine learning methods have gradually emerged. At present, the machine learni ng models used for Chinese named entity recognition mainly include HiddenMa rkovmodel (HMM), conditional randomfield (CRF)[3] and so on. With the impro vement of hardware computing power, the methods based on deep learning are more and more common, and the effect is better than the methods based on statistical machine learning. At present, the methods based on deep learning ma inly train the model through neural networks. The mainstream neural network models include convolutional neural networks (CNN)[4], recurrent neural networ ks (RNN)[5] and so on. In the data relation extraction task in the field of tradi tional Chinese medicine, some scholars have used the pipeline relation extractio n model to extract the relationship of traditional Chinese medicine (TCM) texts. Xie et[6] al. use Long-short term memory (LSTM) network to recognize entiti es from the labeled data, and then classify the extracted entities for relation ex traction to complete the extraction of the entire triplet. In the process of classi fication, through the convolutional neural network (ConvolutionalNeuralNetwork, CNN) entity relationship of polysemy knowledge fusion. Zhang et[7] al. use co nditional random fields for entity recognition and extraction, use crawlers to cr awl entity attributes, and use BiLSTM with attention mechanism for relation ex traction, and realize the processing of polysemy through entity attributes. Wang Shang[8] used a comprehensive cross-entropy loss function and the SEGATT la yer of the segmented attention mechanism for relation classification, and used CNN for knowledge fusion.

By using classical natural language processing methods, this study first segments the text data of classical ancient books, and performs named entity recognition according to the general PKU scheme. On this basis, the TF-IDF algorithm is used to extract key entity words, and then the dependency syntax analysis is carried out to provide data samples for subsequent knowledge graph construction. In the implementation of the specific scheme, this study uses conditional random field natural language processing model +TF-IDF algorithm key entity extraction algorithm + high-performance dependency syntax parser based on neural network to automatically analyze and visualize the representative text data of Yishui School, which provides reference for the research and application of artificial intelligence technology and traditional Chinese medicine school.

## 2. Materials and Methods

### 2.1 Experimental Data

The data of this study are the publicly available versions of the Origin of Medicine, Spleen and Stomach Theory and Yin Syndrome Lue Case. The full text content is transformed into txt documents, the table of contents is removed, only the full text title and all the main text are retained, and Spaces and blank lines are removed, and no data cleaning is performed.

### 2.2 Conditional random fields model

Conditional Random Field (CRF) is a basic model of natural language processing, which is widely used in Chinese word segmentation, named entity recognition, part-of-speech tagging and other tagging scenarios. Chinese word segmentation uses BMES word position method, that is, word head, word middle, word end and independent word. The input sentence S is equivalent to the sequence X, and the output label sequence L is equivalent to the sequence Y. We want to train a model to find the optimal corresponding L under the premise of a given S. The key point of training this model is the selection of the feature function F and the determination of the weight of each feature function W. For each feature function, its input has the following four elements: ① Sentence S②i, which is used to represent the ith word in sentence S ③li, which is the part-of-speech (POS) tagged by the scoring sequence to the ith word ④li−1, which is the part-of-speech tagged by the scoring sequence to the ith word. Its output value is either 0 or 1, where 0 means that the sequence to be scored does not conform to this feature, and 1 means that the sequence to be scored does. For the sequences L and S, we can construct the conditional probability distribution model formula:

$$P(L, S) = p(l_1) \prod_i p(l_i|l_{l-1}) p(w_i|l_i)$$

IOB labeling method is used for named entity recognition on the basis of word segmentation, as shown in Figure 1, using the gram generated by the word segmentation sequence of each sentence, the tri-gram model is used to extract features, and finally input into the CRF model to complete the labeling.

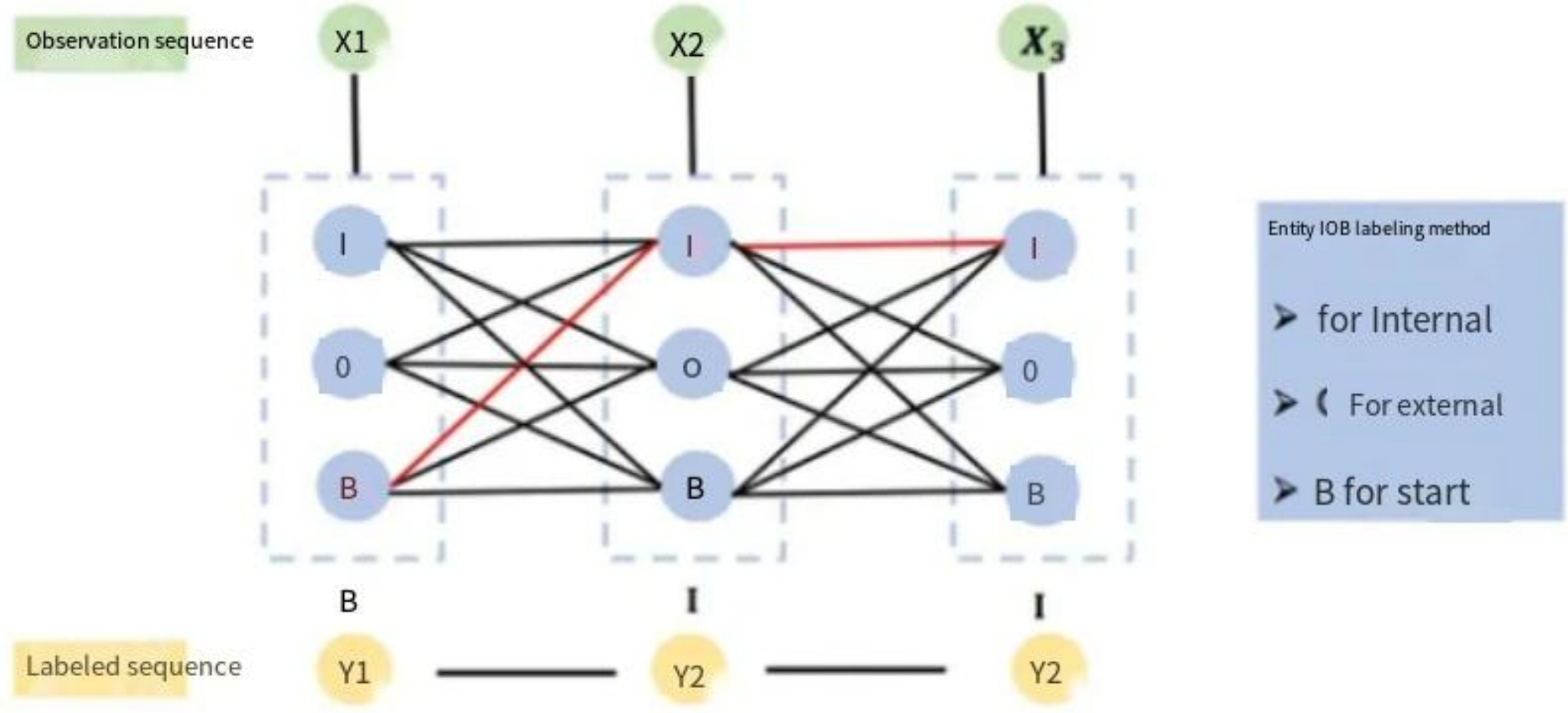

Figure 1 Entity labeling schema diagram of random conditional field model

## 2.3 TF-IDF algorithm

TF-IDF (Term frequency-inverse Document Frequency) is a common weighting technique used in information retrieval and data mining, which is often used to mine keywords in articles. TF-IDF is a statistical analysis method used to evaluate the importance of a term to a document set or a corpus. TermFrequency (TF) is the number or frequency of occurrences of a term in a document. If a term appears more than once in a document, it is likely to be an important term. The formula is as follows:

Term Frequency (TF) = The number of times a term appears in a document/the total number of words in the document

Inverse Document Frequency (IDF) =log(total number of documents in the corpus /(number of documents containing the term +1))

TF−IDF= Term frequency (TF) × Inverse Document frequency (IDF)

The importance of a term is directly proportional to the number of times it appears in the document and inversely proportional to the number of times it appears in the corpus. This calculation method can effectively avoid the influence of common words on keywords, and improve the correlation between keywords and articles.

## 2.4Dependency Parser Based on Neural Network

Dependency parsing can help us understand the meaning of text. Grammar parsing is an important part of language understanding. The goal is to analyze the grammatical structure of a sentence and represent it into an understandable structure, usually a tree structure. The dependency syntax theory believes that there is a master-slave relationship between words. If a word modifies another word in a sentence, the modifier is called a dependent word, and the modified word is called a dominant word.

Dependency parsing based on neural network converts the sequence of words in a sentence into a graph structure by analyzing the grammatical relationship inside the sentence. Common grammatical relations include verb-object relation, left-adjunction relation, right-adjunction relation, coordinate relation, determination-central relation, subject-verb relation, etc. Dependency grammar is a commonly used grammar system. Dependency arc connects two sentences in a sentence that have a certain grammatical relationship, forming a syntactic dependency tree. The dependency tree is constructed by the stack, starting from the root node, and then all the words stored in the cache are pushed into the stack one by one by using three states: Shift, Left-Reduce and Right-Reduce. Neural network consists of three layers, Input layer (Softmax layer), Hidden layer (Hidden layer) and output layer (Input layer). The model is from the 2014 paper "A Fast and Accurate Dependency Parser using Neural Networks" by Danqi Chen and Christopher D. anning. In this study, HanLP is used to implement a dependency parser based on neural networks.

## 2.5 Experimental Environment

The research was implemented on a small artificial intelligence platform equipped with Intel Xeon Gold 6248R CPU@3.00Ghz*96, 256GB memory and NVIDIA A100 80G*2 GPU computing card in the Laboratory of Traditional Chinese Medicine Informatics of Hebei University. Ubuntu 18.04.6LTS, Python 3.9 environment.

# 3 Experimental results

## 3.1 Results of word segmentation and entity recognition

The experiment completed word segmentation and entity recognition of the full texts of the Origin of Medicine, Spleen and Stomach Theory and Yin Syndrome Lue Case, and obtained 472, 899 and 726 corpus items respectively. Due to the difficulty in defining the entity attributes of the text data of ancient Chinese medicine books, this study only divides the entity categories such as nouns, verbs, adjectives, and modal words, and focuses on the meaning of the entity words. The word frequencies and TF-IDF evaluation importance extracted by relevant natural language processing are shown in Table 1-3.

Table 1 Results of natural language processing for the Origin of Medicine

| Corpus | Word frequency | Corpus | Importance |
|---|---|---|---|
| Licorice | 99 | Inner meridian | 0.144594626 |
| Internal meridian | 93 | Licorice | 0.138978263 |
| Preparation | 93 | Urinating | 0.084229251 |
| Cloud injection | 93 | Secret essentials | 0.078901022 |
| Among | 89 | Atractylodes | 0.07506739 |
| Cannot | 61 | Bitter taste | 0.071468601 |
| One or two | 61 | Peeling | 0.067151327 |
| Urinating | 60 | Scutellaria | 0.060755962 |
| Half two | 49 | Windbreaks | 0.058245271 |
| Atractylodes | 48 | Scent | 0.055884185 |
| Smell | 48 | Air chill | 0.054280297 |
| Per suit | 45 | Ginseng | 0.050494625 |
| Its use | 45 | Peony | 0.049870365 |
| Bitter in taste | 44 | Pure Yang | 0.049474438 |
| Above | 43 | Spleen and stomach | 0.048775877 |
| Peeling | 42 | Headache | 0.046924155 |
| The Secret | 41 | Provision | 0.04679336 |
| Even so | 39 | Huang Lian | 0.044396385 |
| Ginseng | 39 | Bupleurum | 0.04327629 |
| Rhubarb | 39 | Virtuori | 0.043049891 |

Table 2 Results of natural language processing of Spleen and Stomach Theory

| Corpus | Word frequency | Corpus | Importance |
|---|---|---|---|
| Spleen and stomach | 121 | Spleen and stomach | 0.19593789 |
| Cannot | 80 | Yang Qi | 0.117905035 |
| Diet | 72 | Licorice | 0.105701878 |
| Five points | 69 | Ginseng | 0.105225239 |
| Yang qi | 68 | Wong Pak | 0.095679668 |
| Ginseng | 68 | Diet | 0.094028705 |
| Roll down | 64 | Yuan Qi | 0.092605142 |
| Licorice | 63 | Five Organs | 0.087256425 |
| Three points | 60 | Urinating | 0.085568187 |
| Two points | 58 | Stomach gas | 0.071585434 |
| Chi | 57 | Cohosh | 0.070785845 |
| Yuan Qi | 54 | Atractylodes | 0.067288841 |
| urinating | 51 | Top piece | 0.0648743 |
| Five Zang organs | 50 | Limbs | 0.064120618 |
| Insufficient | 49 | Huang Lian | 0.063331257 |
| Wong Pak | 49 | Nine OBE | 0.061112835 |
| limbs | 45 | Peony | 0.059603671 |
| Poop | 40 | Orange peel | 0.057272579 |
| Angelica | 40 | Bupleuri | 0.057073232 |
| Per serving | 40 | Alisma | 0.055762027 |

Table 3 Results of natural language processing for "Yin Syndrome Lue"

| Corpus | Word frequency | Corpus | Importance |
|---|---|---|---|
| Yin syndrome | 104 | Yin Syndrome | 0.272813995 |
| Yin Yang | 63 | Shaoyin | 0.135025106 |
| Hand and foot | 62 | Venomous | 0.12329094 |
| Typhoid fever | 61 | Zhong Jing | 0.115980516 |
| Shaoyin | 55 | Yin and Yang | 0.107359649 |
| Four Inverse soup | 53 | Yang Qi | 0.105655059 |
| Limbs | 50 | Aconite | 0.104053931 |
| Yang Qi | 50 | Limbs | 0.086826449 |
| One or two | 47 | Dried ginger | 0.078298201 |
| Venomous | 47 | Yang Certificate | 0.073449922 |
| Aconite | 45 | Yin Qi | 0.072152558 |
| Half two | 42 | The Lunar Moon | 0.072066233 |
| Do not | 41 | Atractylodes | 0.070615346 |
| Cannot | 38 | Aquifer clouds | 0.068157169 |
| Zhong Jing | 37 | Licorice | 0.067476562 |
| fever. | 37 | Urinating | 0.067476562 |
| Irritability | 35 | Internal injuries | 0.065550539 |
| Living people | 34 | Yin vein | 0.062704596 |
| Yin Qi | 34 | Living People | 0.061867588 |
| Dried ginger | 33 | Chung Jing Yun | 0.059978309 |

3.2 Visualization results of related entity vocabulary map

The relevant data were collated and summarized, and the word cloud map was drawn according to the entity word importance data, as shown in Figure 2.

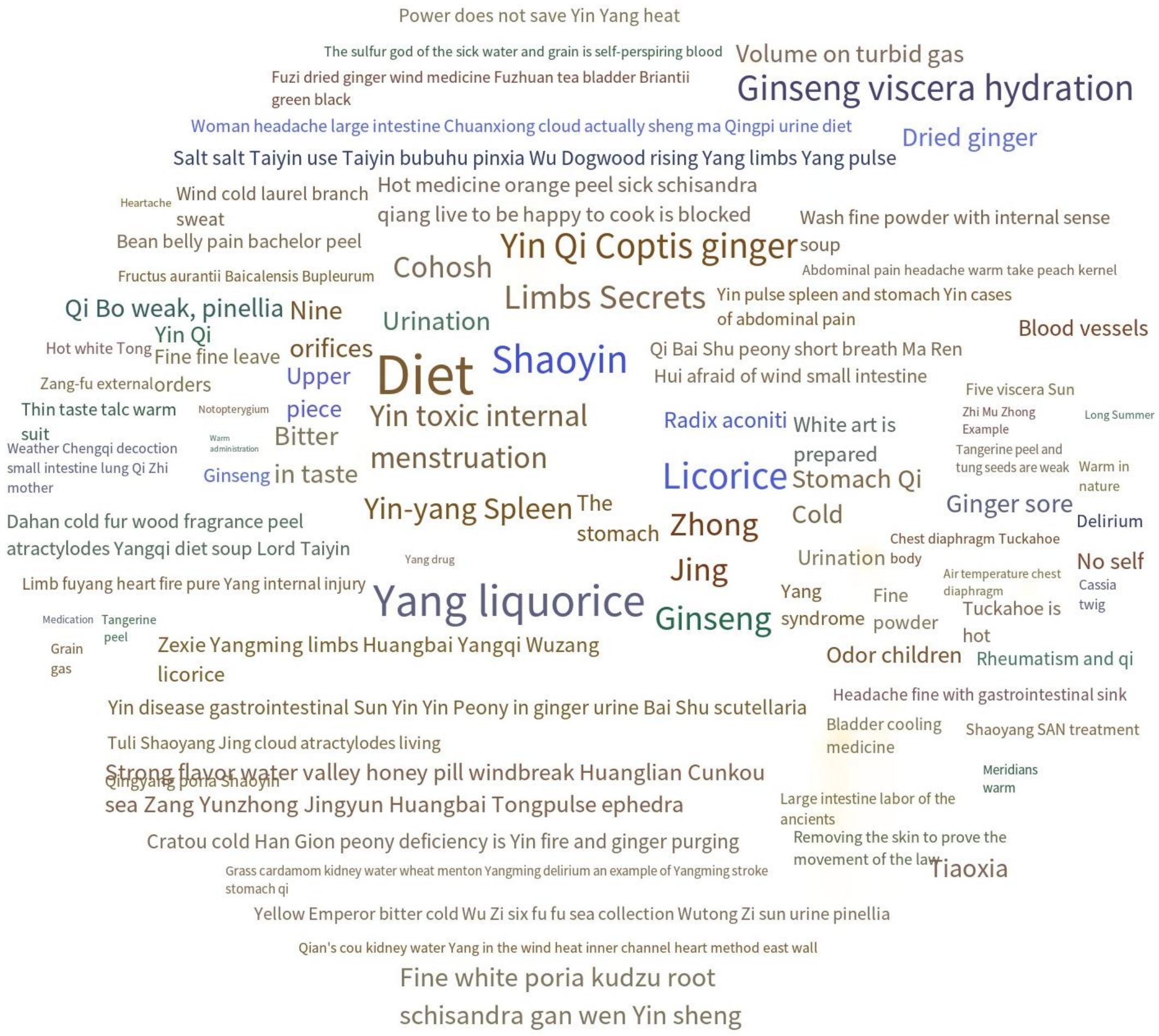

FIG. 2 Word cloud map of key entities in representative works of Yi-shui School

### 3.3 Results of dependency parsing

This study completed the partial syntactic analysis of all the clauses of the three works. Taking the text description of the theory of quoting classics in Medical Qi Yuan as an example, the sample data were extracted for relation extraction and image rendering.

The sample texts are as follows:

*Each sutra quotes the Sun Sutra, Qiang Huo; In the lower yellow cypress, small intestine, bladder also. Shaoyang meridian, Bupleurum; In the lower Qingpi, bile, sanjiao also. Yangming meridian, cohosh, angelica dahurica; In the lower, gypsum, stomach, large intestine also. Taiyin meridian, Baishao medicine,*

*spleen, lung also. Shaoyin meridian, anemarrhena, heart and kidney. Jieyin meridian, Qingpi; In the lower, bupleurum, liver, envelop also. The medicine of the above 12 classics is also.*

The dependency grammar tree is constructed as shown in Figure 3. The model can recognize this text in classical Chinese, analyze its grammatical structure according to the entity recognition results, and extract the relationship between entities. Taking the Sun Meridian as an example, it can clearly distinguish the relationship between the Sun Meridian and Qiang Huo, and between the yellow cypress and small intestine, bladder.

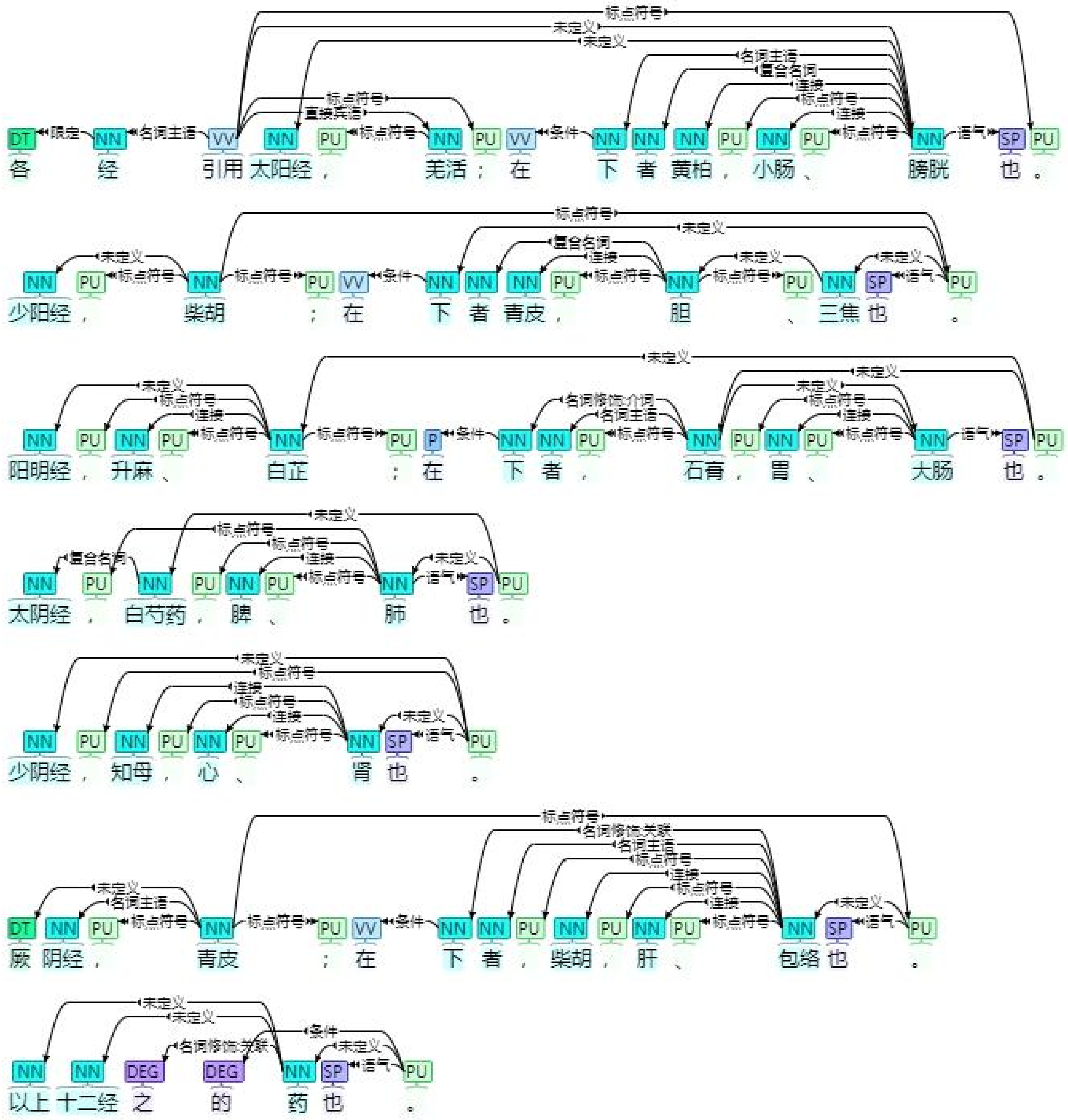

FIG. 3 Dependency grammar tree of quotations from Chinese herbal medicine

## 4 Final Remarks

In this study, the named entity recognition method based on conditional random fields is used to analyze the entity vocabulary, semantic features and syntactic structure of the text data of the Yishui School of traditional Chinese Medicine. The extraction of key named entities from unstructured text data has achieved good results. It has important theoretical and practical guiding value for the summary of academic views of different doctors of the Yishui School, the discovery of differences in academic ideas, and the study of the inheritance of the Yishui School. In the next step, on the basis of named entity recognition, we will continue to study TCM entity relation extraction from classical Chinese data, and then construct the knowledge graph of Yishui School, which provides reference for the application of artificial intelligence methods in the research of TCM school.